%% file: root.tex
\documentclass[letterpaper, 10 pt, conference]{IEEEtran}  

\IEEEoverridecommandlockouts                              

\usepackage[numbers]{natbib}
\usepackage[utf8]{inputenc} 
\usepackage[T1]{fontenc}    
\usepackage{hyperref}       
\usepackage{url}            
\usepackage{booktabs}       
\usepackage{amsfonts}       
\usepackage{nicefrac}       
\usepackage{microtype}      
\usepackage{graphicx}
\usepackage{amsmath,amssymb,amsfonts}
\usepackage[acronym]{glossaries}
\usepackage{comment}
\usepackage{xcolor}
\usepackage[capitalise]{cleveref}
\usepackage{caption}
\usepackage{svg}
\usepackage{listings}
\usepackage{siunitx}
\usepackage[section]{placeins}
\usepackage[english]{babel}
\usepackage{todonotes}
\usepackage{booktabs}
\usepackage{multirow}

\input{acr/acr.tex}

\title{\LARGE \bf
TC-Driver: Trajectory Conditioned Driving for Robust Autonomous Racing - A Reinforcement Learning Approach
}

\author{Edoardo Ghignone$^{1}$, Nicolas Baumann$^{1}$, Mike Boss$^{2}$ and Michele Magno$^{1}$
\thanks{$^{1}$ Associated with Center for Project Based Learning, D-ITET, ETH Zurich}
\thanks{$^{2}$ Associated with D-INFK, ETH Zurich}%
}

\begin{document}

\maketitle
\thispagestyle{empty}
\pagestyle{empty}

\begin{abstract}

Autonomous racing is becoming popular for academic and industry researchers as a test for general autonomous driving by pushing perception, planning, and control algorithms to their limits. While traditional control methods such as  \gls{mpc} are capable of generating an optimal control sequence at the edge of the vehicles' physical controllability, these methods are sensitive to the accuracy of the modeling parameters. 
This paper presents TC-Driver, a \gls{rl} approach for robust control in autonomous racing. In particular, the TC-Driver agent is conditioned by a trajectory generated by any arbitrary traditional high-level planner. The proposed TC-Driver addresses the tire parameter modeling inaccuracies by exploiting the heuristic nature of \gls{rl} while leveraging the reliability of traditional planning methods in a hierarchical control structure. We train the agent under varying tire conditions, allowing it to generalize to different model parameters, aiming to increase the racing capabilities of the system in practice. 
The proposed \gls{rl} method outperforms a non-learning-based \gls{mpc} with a 2.7 lower crash ratio in a model mismatch setting, underlining robustness to parameter discrepancies. In addition, the average \gls{rl} inference duration is 0.25 ms compared to the average \gls{mpc} solving time of 11.5 ms, yielding a nearly 40-fold speedup, allowing for complex control deployment in computationally constrained devices.
Lastly, we show that the frequently utilized end-to-end \gls{rl} architecture, as a control policy directly learned from sensory input, is not well suited to model mismatch robustness nor track generalization. Our realistic simulations show that TC-Driver achieves a 6.7 and 3-fold lower crash ratio under model mismatch and track generalization settings, while simultaneously achieving lower lap times than an end-to-end approach,  demonstrating the viability of TC-driver to robust autonomous racing.

\end{abstract}

\section{INTRODUCTION}

Autonomous car racing pushes the boundaries of algorithmic design and implementation in perception, planning, and control \cite{amz_fullstack}. Thus, it is a valuable asset for researchers to push the limits of autonomous driving \cite{michele_list0, michele_list1}. This offers many benefits, such as enhancing road safety, reducing carbon emissions, transporting the mobility-impaired, and reducing driving-related stress \cite{drivingstress, av_survey}. Therefore, autonomous car racing serves as a catalyst for the long-term goals of general autonomous vehicles \cite{liniger_mpcc}.

In recent years, many autonomous racing competitions have emerged, held at prestigious robotics conferences, and received considerable attention in the fields of robotics and \gls{ai}. Among other competitions, the F1TENTH racing platform is gaining popularity and attracting researchers from all over the globe. F1TENTH is a semi-regular autonomous racing competition involving a race car on a scale of 1:10. The competition offers both a simulator environment \cite{okelly2020f1tenth} and a physical racing platform.
As the standardized platform offers little room for improvement on the hardware side, the main challenges are raised on the algorithmic side \cite{okelly2020f1tenth}.
Namely, the control layer becomes the key focus of development, as the system in itself is highly nonlinear and the behavior of the car must be taken into consideration at the edge of stability \cite{liniger_mpcc}. 

Current \gls{sota} racing controllers utilize optimal control methods such as \gls{mpc} \cite{amz_fullstack, liniger_mpcc, michele_list0, michele_list1}. 
While \gls{mpc} can guarantee optimality of the planned trajectory and tracking within its receding horizon, it heavily relies on the accuracy of the modeling parameters. 
Particularly in the context of autonomous car racing, the model inaccuracies of the lateral tire forces are critical for high-performance racing.
These forces are notoriously difficult to model, and the tires' behavior is highly nonlinear \cite{story_of_modelmismatch}, making modeling errors potentially very dangerous. In real racing scenarios, a tire modeling mismatch is very likely to occur, as high wear and tear and changes in weight modify the initial parameters \cite{story_of_modelmismatch}. While there exist several previous works that have attempted to address this issue using learning-based methods \cite{frohlich2021model,jain_bayesrace_2020} for \gls{mpc}, we assess and highlight the feasibility and performance of an \gls{rl} approach.


\gls{rl} methods offer a \gls{ml}-based solution to handle these mismatches. Said kind of techniques proved to be able to handle high-performance racing on different occasions \cite{chisari2021learning, brunnbauer_model-based_2021, fuchs2021, sophy}. The mentioned architectures are end-to-end learned, meaning that they learn the optimal control policy directly from sensory input.
Further, these works do not focus on the generalisation capabilities on model mismatch and only show partial \cite{brunnbauer_model-based_2021} or poor \cite{fuchs2021} generalisation results to unseen tracks.

We propose a hybrid architecture that specifically addresses the robustness towards model mismatch and track generalization, inspired by the two-layer planner-controller separation that is often present in robotic systems \cite{ar_survey, amz_fullstack}.
In this framework, the planning layer (e.g., Frenet planner \cite{frenetplanner}) is responsible for generating a safe and performant trajectory, while the control layer is dedicated to generating control inputs in order to make the system follow the given trajectory.
According to this layout, we consider the planner to be given, and use a \gls{rl} agent for the low-level control, exploiting the learning capabilities of such models to heuristically handle model mismatch and track generalization, and leverage the safety and reliability of traditional planning methods \cite{frenetplanner}. 
Namely, we present a trajectory-conditioned \gls{rl} controller (TC-Driver) that is able to generalize to different tracks as its task is to track an arbitrary trajectory.

Furthermore, we train the \gls{rl} agent under constantly varying tire parameters, such that it learns to generalize to varying model conditions, allowing lap completion in a racing setting. 
We show that such a hierarchical structure is beneficial to the learning objective of a \gls{rl} agent when compared to an end-to-end setting based on previous \gls{sota} \cite{fuchs2021, song_autonomous_2021, brunnbauer_model-based_2021, chisari2021learning}, as later shown in \cref{sec:results}.

The proposed architecture offers multiple benefits:
\begin{itemize}
    \item \textbf{Robustness to Modeling Mismatch:} The proposed controller adapts to parameter mismatches due to deep learning's generalization capabilities and \gls{rl}'s heuristic properties. In particular, we focus on the notoriously difficult to model lateral tire forces required for high-speed racing \cite{liniger_mpcc,frohlich2021model,jain_bayesrace_2020,story_of_modelmismatch}. 
    It is shown in \cref{tab:robustness} that the proposed method yields better model mismatch robustness compared to the non-learning-based \gls{mpc} and the end-to-end setting. The \gls{mpc}, under model mismatch, demonstrates a crash ratio of 38.1\% as opposed to the TC-Driver with 14.29\%, resulting in a factor of $\sim2.7$ lower crash ratio. Further, the end-to-end architecture demonstrates a crash ratio of 95.24\%, a lower crash ratio of TC-Driver is demonstrated by the factor of $\sim6.7$. Thus underlining the model mismatch robustness.

    \item \textbf{Track Generalization Capabilities:} The proposed architecture can better generalize to unseen tracks as the observation given to the \gls{rl} model has no general reference to the track itself but only a partial trajectory.
    The proposed architecture yields superior generalization capabilities on unforeseen tracks when compared to the end-to-end setting based on previous \gls{sota} \cite{fuchs2021, chisari2021learning, song_autonomous_2021}.
    As shown in \cref{tab:generalize} the average crash ratio of the end-to-end architecture is 38.09\%, while the TC-Driver's crash ratio is 12.7\%, demonstrating a lower crash ratio by the factor of $3$.
    When compared to \gls{mpc}, our method does not show better results. However, on two tracks out of three, it yields the same performance as the optimal control method.

    
    \item \textbf{Computational Benefit:} The envisioned control structure is beneficial in terms of computational load compared to a full \gls{mpc} setting. To quantify, the \gls{rl} inference has an average duration of \SI{0.25}{ms} compared to the average \gls{mpc} solving time of \SI{11.5}{ms}, as in \cref{tab:comptime}.
    
\end{itemize}

\section{METHODOLOGY}
The \gls{rl} terminology follows the convention of \cite{Sutton1998}. The main goal of our architecture is to train an agent operating a race car that is aware of a given trajectory under the influence of noise applied to the tire friction coefficients. That is, in every episode, the environment will have different tire modeling parameters. Thus, the agent learns to handle the tire parameter modeling mismatch as it generalizes on how to handle these conditions during training, ultimately allowing for robust tracking of a given trajectory.

\subsection{Simulation Environment}\label{subsec:sim}
The F1TENTH simulation environment \cite{okelly2020f1tenth} aims to offer an OpenAI gym compatible wrapper \cite{openai}. 
Within the environment, the vehicle's dynamics are modeled with the \emph{Single Track} model \cite{althoff_commonroad_2017}, which is also known as the \emph{Bicycle Model}, to realistically simulate Ackermann-steered vehicles. 

\begin{figure}[!h]
    \centering
    \includegraphics[width=0.85\columnwidth]{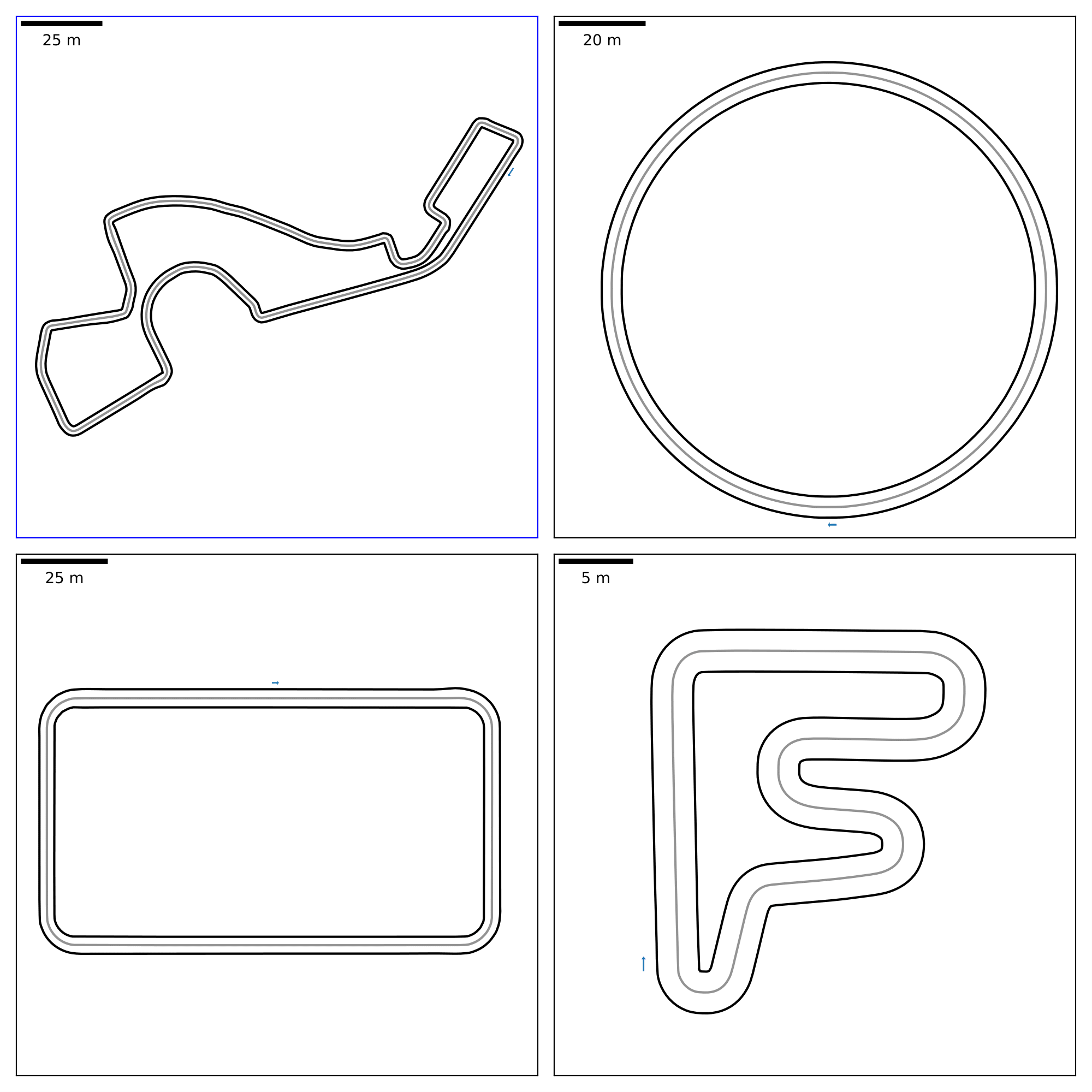}
    \caption{Training track \emph{SOCHI} depicted in blue. Testing tracks \emph{Circle}, \emph{Square} and \emph{F}, which are unseen during training. The tracks vary in length from \SI{470}{m} to \SI{89}{m} in length, while the gray line depicts the centerline. Track width varies from \SI{5}{m} to \SI{3}{m}, for reference the width of the car is \SI{0.31}{m}.}
    \label{fig:tracks}
\end{figure}

$\mu,\,C_{S,\,f},\,C_{S,\,r}$ model the friction, the cornering stiffness on the front axle, and the cornering stiffness on the rear axle, respectively as in \cite{bicycle_model, althoff_commonroad_2017}. The F1TENTH environment has been modified to be able to inject noise into the parameters, allowing the investigation of robustness in terms of tire modeling inaccuracies. The simulation environment offers the following car's dynamic state: 
$s_{dyn} = [s_x,\,s_y,\,\psi,\,v_x,\,v_y,\,\dot{\psi}]$ = [global x position, global y position, yaw angle with respect to the positive x-axis, longitudinal velocity, lateral velocity, yaw rate]. 

Further, the simulation environment provides sensory input in the form of a LiDAR scan made of 1080 points over $270^{\circ}$ coverage area around the car. To summarize, the observation of the environment is $obs_{gym} = [scan, \,s_x,\,s_y,\,\psi,\,v_x,\,v_y,\,\dot{\psi}]$

The action space of the gym environment solely consists of continuous actions $a = [v,\, \delta]$, where $v$ is the desired longitudinal velocity and $\delta$ is the steering angle of the agent.

The reward function is defined in \cref{eq:basereward} inspired by \cite{chisari2021learning, fuchs2021}. 

\begin{equation}\label{eq:basereward}
r_{t}= \begin{cases}-c & \text { if track constraints are violated } \\ p_{t+1}-p_{t}& \text { otherwise }\end{cases}
\end{equation}

where $c = 0.01$ and $p_{t+1} - p_t$ is the track advancement within one simulation time step. The range of the reward is calculated as $p_{t+1} - p_t \in [-T_s V_{max}, T_s V_{max}]$ and evaluates to $p_{t+1} - p_t \in [0, 0.1]$ with the simulation time step $T_s = 10\,ms$ and $V_{max} = 10 \frac{m}{s}$. The track constraints are defined as $|n_t| \ge \frac{1}{2} w_{track} - \frac{3}{2} w_{car}$. If the car deviates from the center line by more than half of the track's width $w_{track}$ minus a safety margin based on $\frac{3}{2}$ times the car's width $w_{car}$, it violates the track constraints. This emulates on the \gls{rl} the spatial soft constraints also employed on the \gls{mpc} as defined in \cite{liniger_mpcc}.

\subsection{Reinforcement Learning Architectures}
In this section, we introduce both the frequently used end-to-end \gls{rl} architecture \cite{chisari2021learning, sophy, fuchs2021, brunnbauer_model-based_2021} and the low-level \gls{rl} trajectory tracker, with their underlying architecture, environment interaction, and hyperparameters. To ensure a fair comparison, both settings use the same underlying model-free architecture, namely \gls{sac} \cite{sac} featuring off-policy learning, entropy regularization, and double learning. It is derived from the \gls{sb3} \cite{stable-baselines3} implementation.

\subsubsection{End-to-End Racer}

To generate a baseline for comparisons, we utilized the frequently used model-free end-to-end architecture of \cite{chisari2021learning, song_autonomous_2021, fuchs2021}, possessing a slightly modified observation space with respect to the one previously defined in \cref{subsec:sim}. 
The chosen observation space recasts the observation in a \emph{Frenet frame}, which is a representation relative to a trajectory, as in \cite{chisari2021learning, song_autonomous_2021, fuchs2021}. 
The new observation is $obs_{end2end} = [p, n, \psi, v_x, v_y, \dot{\psi}]$ = [progress along the path, perpendicular deviation from the path, yaw relative to the trajectory, relative heading, longitudinal velocity, lateral velocity, yaw rate].
 
 The agent learns a control policy with online environment interaction based on the previously defined reward function in \cref{subsec:sim}. 
 Since advancement-based rewards like ours were broadly tested \cite{fuchs2021, chisari2021learning, brunnbauer_model-based_2021, song_autonomous_2021}, we consider this agent a reasonable comparison model. 
 
 \subsubsection{Trajectory Conditioned Driver}
The proposed low-level trajectory tracker (TC-Driver) tracks the spatial trajectory generated by a high-level planner. Within this work, a pre-generated \gls{mpcc} trajectory is used, which has been custom implemented for this task following \cite{liniger_mpcc}. That is, the track has already been traversed by a \gls{mpc}, and the logged trajectory can then be used by subsequent \gls{rl} agents. It is worth mentioning that this trajectory could be chosen arbitrarily, such as, for example, by using the center-line trajectory instead of the time-optimal \gls{mpc} trajectory. 

The observation space of the proposed low-level trajectory tracker is altered compared with the end-to-end setting. To enable trajectory following, we add a sample of the optimal trajectory relative to the current position of the car. This sample consists of 30 points, rotated and translated to be in the car's frame of reference.

Therefore, the observation space in the hierarchical setting is newly defined as $obs_{traj} = [traj, p,n, \psi,v_x,v_y, \dot{psi}]$ = [relative trajectory, progress along the path, perpendicular deviation from the path, relative heading, longitudinal velocity, lateral velocity, yaw rate].

Furthermore, the reward function is slightly modified to penalize the agent for excessive swerving from the trajectory:

\begin{equation}
r_{t}= \begin{cases}-c & \text { if constraints violated } \\ p_{t+1}-p_{t} - |n_t|& \text { otherwise }\end{cases}
\end{equation}

where $|n_t|$ is the spatial deviation perpendicular to the target trajectory of the planner and the constraint violations are the same as in \cref{eq:basereward}.

\subsection{Tire Parameter Randomization} \label{subsec:tirerand}
The F1TENTH simulation environment utilizes the single-track dynamic model of \cite{althoff_commonroad_2017}.
To apply randomness to the tire coefficients, we added Gaussian noise at each reset of the gym environment during training. The noise was centered at the nominal value friction, used in the \gls{mpc} to find the optimal trajectory.

To find the standard deviation, the limit of tire friction at which \gls{mpc} would not be able to correctly complete a lap was analyzed. Then the standard deviation of the noise was set to be half of that value for the noise to be mostly (but not entirely) inside the range of values that allow \gls{mpc} to finish a lap. The numerical values are: $\mu_{noisy} \sim \mathcal{N}(1.0489,\, 0.0375)$.
 
 \subsection{Implementation}
The used environment is based on an adapted version of the  F1TENTH gym racing environment \cite{okelly2020f1tenth}. Both \gls{rl} agents were implemented using \gls{sb3} \gls{sac} algorithm. \gls{sac} was initialized with gamma at 0.99, an episode length of 10000, batch size of 64, train frequency of 1, and using the \gls{mlp} policy.


\section{RESULTS}\label{sec:results}
In this section, we evaluate the proposed trajectory tracking agent against the end-to-end agent as well as the \gls{mpc} method with and without the tire parameter randomization during training. Evaluation metrics consist of lap-time, the ability to handle different track conditions, and the ability to drive unseen tracks. In simulation, one can use the optimal trajectory of a \gls{mpc} with exact model parameters (without \emph{tire noise}) as the ground truth reference.
For the results presented in \cref{subsec:comp} and \cref{subsec:rob}, an agent was trained for $1e6$ timesteps for each algorithm on the track named \lstinline{SOCHI}, which can be seen in \cref{fig:tracks}.

\begin{figure}[!h]
    \centering
    \includegraphics[width=0.95\columnwidth]{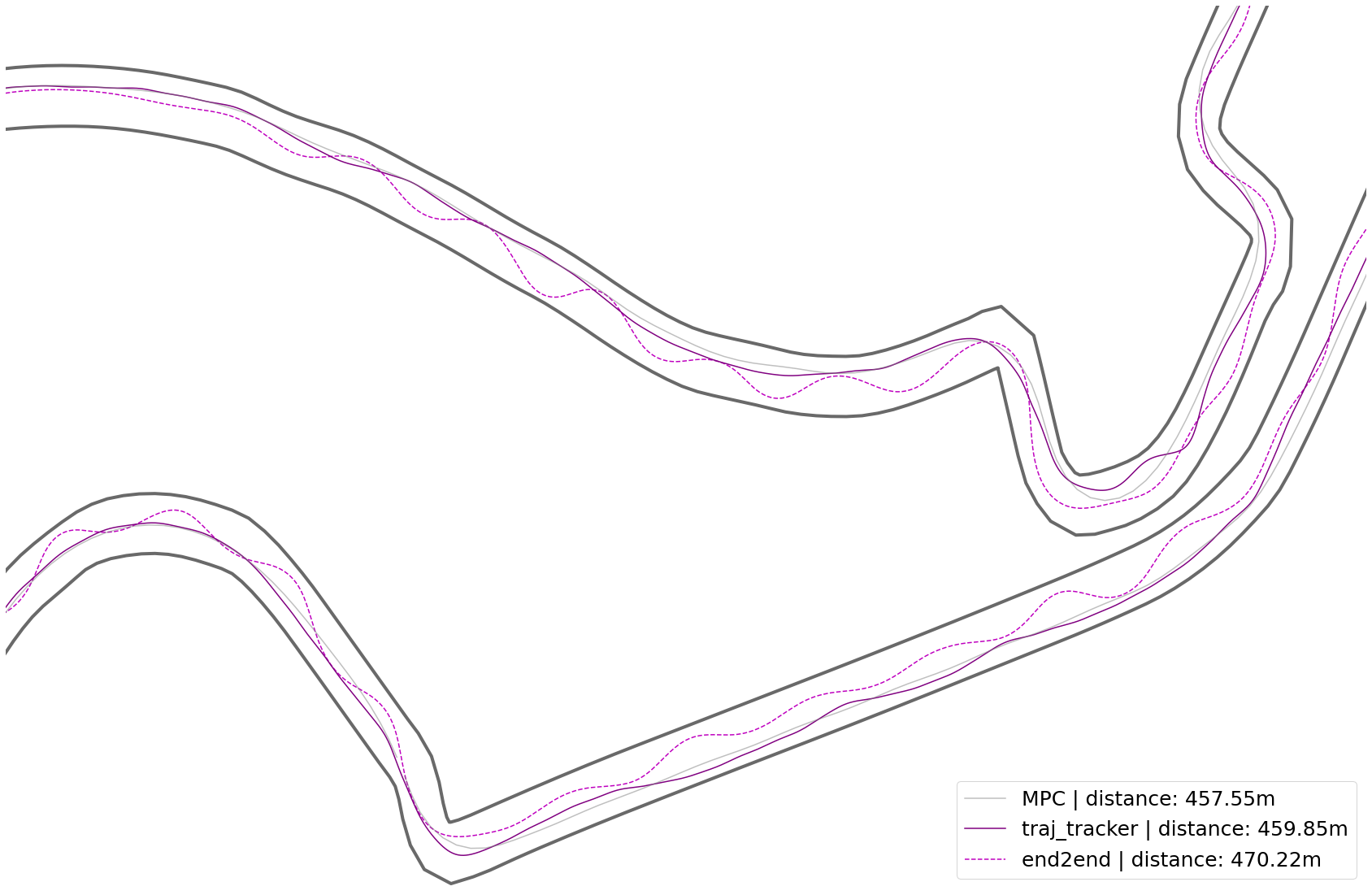}
    \caption{Comparison of the end-to-end and trajectory tracking agent with the optimal trajectory of a \gls{mpc}.}
    \label{fig:trajvsend}
\end{figure}

\subsection{Comparison of End-to-End and TC-Driver}\label{subsec:comp}
It is to be mentioned that a direct comparison between the end-to-end and the trajectory tracker based on the reward itself is not possible as both settings have different reward functions and observation spaces. As a result, we compare the traversed distances of the proposed trajectory tracking agent and the end-to-end agent to that of a \gls{mpc} with exact modeling and simulation parameter matching. \cref{fig:trajvsend} shows that the end-to-end agent swerves much more compared to the trajectory tracker that mostly tracks the \gls{mpc} generated trajectory in a stable manner. Analysing the total run, the end-to-end agent chooses a trajectory that totals 470.22 m in length, which is significantly longer than the proposed TC-Driver trajectory; namely, 12.67 m compared to the 2.3 m of excessive trajectory length of the trajectory tracking agent. This behavior clearly does not follow a realistic purpose as the end-to-end agent swerves a lot on the straights for which the optimal trajectory given by the \gls{mpc} is close to completely straight. 

\begin{figure}[h]
    \centering
    \includegraphics[width=\columnwidth]{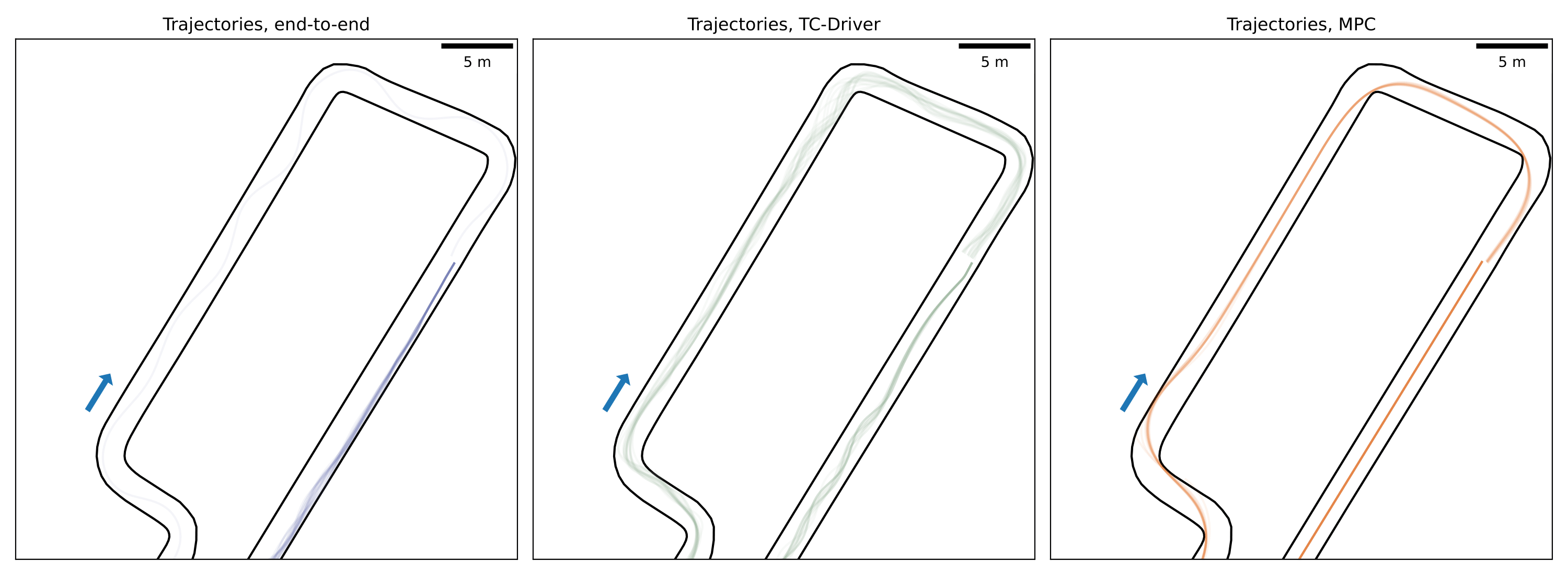}
    \caption{Agents that were trained under tire friction randomization within the \gls{mpc} tolerance for 21 runs, tested in an environment outside the trained tire friction domain. Left is the end-to-end agent; middle shows the proposed trajectory tracking agent; right shows \gls{mpc} with a crash rate of 38.1\% due to the high parameter mismatch.}
    \label{fig:robustness}
\end{figure}

\subsection{Robustness to Tire modeling Mismatch}\label{subsec:rob}
To test the capabilities of the algorithms to generalize to different tire friction, 21 randomly extracted values were utilized during test laps.
To better test the generalization capabilities, these friction parameters were extracted in an interval that was predominantly outside the training range.
In detail, the normal distribution had a mean $0.2$ lower than the nominal one, with the same standard deviation as in the training phase, i.e., $0.0375$.
The \gls{mpc} was run with the nominal system model, i.e., the tires' friction was not changed, to simulate model mismatch. 
The three different models were run on the track, starting from the same position, for one lap. 

In \cref{fig:robustness} one can see a trajectory extract, with the 21 laps superimposed one on the other. 
Due to the parameters mismatch, the \gls{mpc} does not manage to finish the lap in 38.1\% of the times, as shown in \cref{tab:robustness}. Yet, it still achieves the best and most consistent lap time of \SI{46.261}{s} with \SI{0.054}{s} standard deviation. While the trajectory conditioned \gls{rl} driver is significantly slower with an average lap time of \SI{53.528}{s}, it has the lowest crash percentage of 14.29\%. Compared to the end-to-end \gls{rl} driver with a crash ratio of 95.24\% (only a single run was completed successfully), the proposed trajectory conditioned algorithm performs notably better when faced with model mismatch. 

Investigating the crash ratio of \cref{tab:robustness}, the proposed TC-Driver outperforms the end-to-end architecture by a factor of $\sim 6.7$ and the \gls{mpc} by a factor of $\sim 2.7$. In this setting, the proposed method shows better generalization capabilities to model mismatch. Regarding the \gls{mpc} it has to be said that such a result is expected, as the tire mismatch lies outside of the modeling domain. A solution for such a situation would be the integration of learnable parameters within the \gls{mpc} model, as in \cite{jain_bayesrace_2020}. Thus, this result does not exhibit superiority to \gls{mpc}, but rather demonstrates a case in which \gls{rl} can be utilized in the mitigation of model mismatch.

\begin{table}[h!]
\captionof{table}{Lap time results of 21 runs, comparison with imperfect knowledge of dynamics on the training track. Average lap time $t_{\mu}$ in seconds (lower is better); Standard deviation of the lap times $t_{\sigma}$ (lower is better); Percentage of crashes during the runs (lower is better).}\label{tab:robustness}
    \resizebox{\columnwidth}{!}{\begin{tabular}{@{}llll@{}}
    \hline
    & Lap time $t_{\mu}$ [s] & Lap time $t_{\sigma}$ [s] &   Crashes   \\ \hline\hline
    MPC &  {\bf 46.261  }        &  {\bf 0.054 }       &   38.10\%   \\ \hline
    end-to-end  &   59.030        &     n.a.         &   95.24\%  \\ \hline
    TC-Driver  &    53.528     &    0.348       &   {\bf 14.29 }\%    \\ \hline
    \end{tabular}}
\end{table}

\subsection{Track Generalization Capabilities}
To test the trajectory conditioned driver's ability to generalize to race tracks beyond the training track of \emph{SOCHI}, it was evaluated on three additional unforeseen tracks, namely \emph{Circle}, \emph{Square} and \emph{F} track visible in \cref{fig:tracks}.  

The agents were started at 21 different positions along these tracks and drove a single lap each. To further emphasize the generalization capabilities to arbitrary trajectories, the trajectories used for conditioning the TC-Driver were the centerlines of the test tracks, instead of the time-optimal raceline of the \gls{mpcc} as in the training phase.

\cref{tab:generalize} depicts the described runs on the unseen tracks. The \gls{mpc} clearly outperforms both \gls{rl} methods, as expected. It shows the fastest lap times on all test tracks with the lowest standard deviation, while never crashing. Averaging the lap time standard deviation and comparing them between the end-to-end and TC-Driver, yields a factor $\sim3$ smaller deviation in favor of the TC-Driver. Thus, the TC-Driver manages to complete the laps in a significantly more consistent time. Further, evaluating the average crash ratio of the end-to-end (38.09\%) and TC-Driver (12.7\%) results in a factor of $3$ lower crash ratio for the TC-Driver. Therefore, the TC-Driver shows better generalization capabilities on unseen tracks. 

Interestingly, however, is the inferior crash ratio of the TC-Driver on the \emph{F}-track. The suspected reason for this result is the scale of the track with respect to the training track \emph{SOCHI} and the other test tracks \emph{Circle} and \emph{Square}.

\begin{figure}[h!]
    \centering
    \includegraphics[width=\columnwidth]{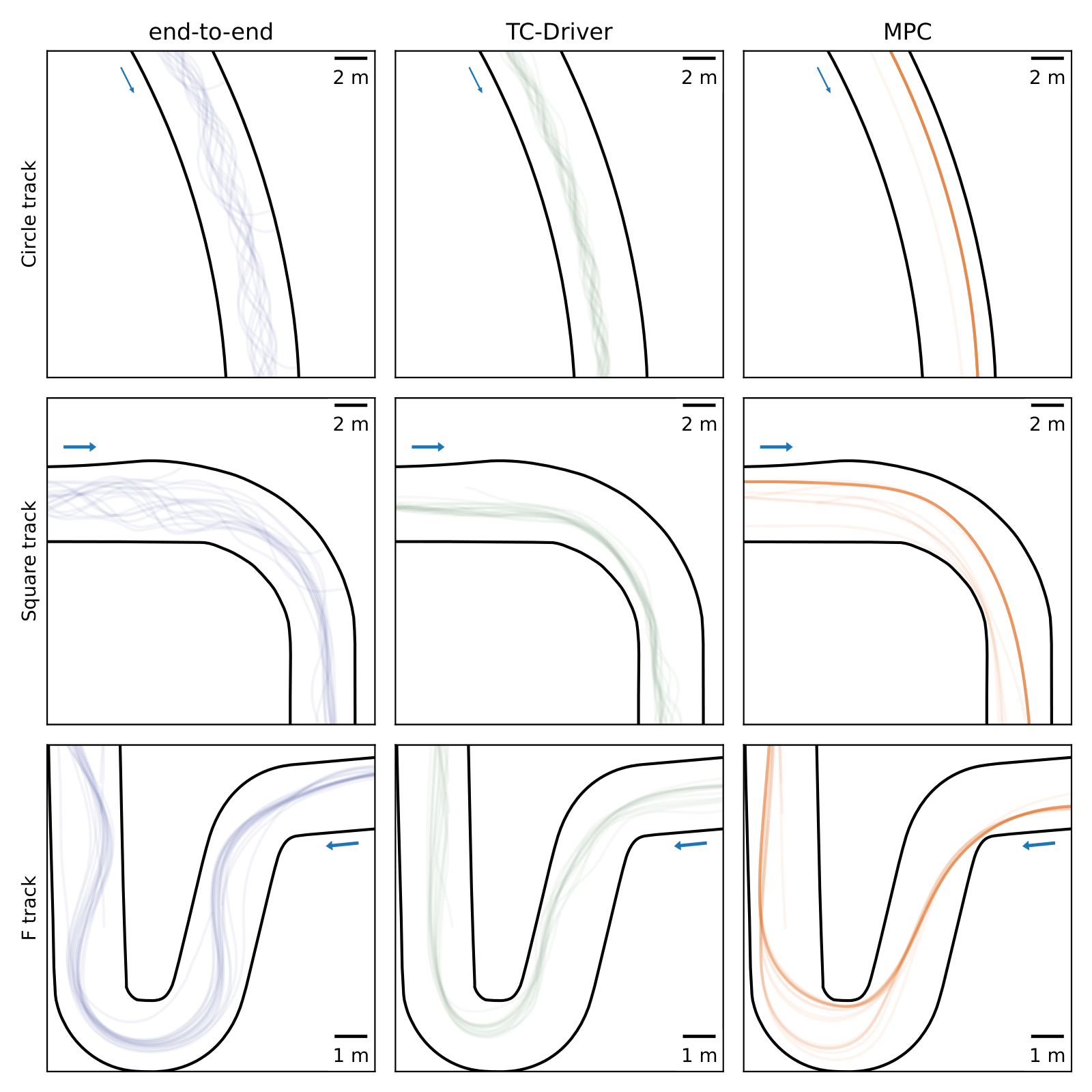}
    \caption{Generalization runs of the end-to-end, TC-driver and \gls{mpc} algorithms from left to right, on the unseen test tracks \emph{Circle}, \emph{Square} and \emph{F} respectively, from top to bottom. Consisting of 21 runs without tire friction randomization.}
    \label{fig:gen}
\end{figure}

\begin{table}[h]
\centering
\captionof{table}{Lap time results of 21 runs on the unseen tracks \emph{Circle}, \emph{Square} and \emph{F} with zero model mismatch.  Average lap time in seconds (lower is better); Standard deviation of the lap times (lower is better); Percentage of crashes during the runs (lower is better).}\label{tab:generalize}
\resizebox{\columnwidth}{!}
{\begin{tabular}{@{}lllll@{}}
  \toprule
  Track & Driver & Lap time $t_{\mu}$ [s] & Lap time $t_{\sigma}$ [s] &   Crashes \\
  \midrule
  \multirow{3}{*}{\emph{Circle}} 
  & \gls{mpc} &  {\bf 33.111  }        &  {\bf 0.129 }       &  0.00\%   \\ 
  & end-to-end  &   39.436        &     0.456         &   61.90\%  \\ 
  & TC-Driver  &    39.548     &    0.335       &   {\bf 0.00 }\%    \\ 
  \midrule
  \multirow{3}{*}{\emph{Square}} 
  & \gls{mpc} &  {\bf 39.544  }        &  {\bf 0.251 }       &  {\bf 0.00\%}   \\ 
  & end-to-end  &   49.107        &     0.605         &   28.57\%  \\ 
  & TC-Driver  &    46.992     &    0.333       &    {\bf 0.00\%}    \\ 
  \midrule
  \multirow{3}{*}{\emph{F}} 
  & \gls{mpc} &  {\bf 10.582  }        &  {\bf 0.210 }       &  {\bf 0.00\%}   \\ 
  & end-to-end  &   14.303        &     1.750         &   23.81\%  \\ 
  & TC-Driver  &    11.753     &    0.274       &    38.10\% \\
  \bottomrule
\end{tabular}}
\end{table}

\subsection{Computation Time}
Lastly, we focus on the computational time of the utilized control methods. \cref{tab:comptime} depicts the average computation time of each method and their respective standard deviation. The \gls{mpc}'s average computation duration is approximately \SI{11}{ms} with a rather high standard deviation of \SI{0.9}{ms}. The reason for the higher deviation arises from the nature of quadratic programming, which is subject to constantly varying solving conditions. On the other hand, both \gls{rl} algorithms show a significantly lower and more constant inference time of approximately \SI{0.26}{ms}. Thus, the \gls{rl} computation time is faster by a factor of roughly 40.

\begin{table}[h]
\captionof{table}{Average computation time of the utilised control methods and their respective standard deviation.}\label{tab:comptime}
\centering
\resizebox{\columnwidth}{!}
{\begin{tabular}{@{}lll@{}}
\hline
& Computation Time $t_{\mu}$ [ms] & Computation Time $t_{\sigma}$ [ms] \\ \hline\hline
\gls{mpc}   &  11.2     &   0.9  \\ \hline
end-to-end  &    {\bf 0.26 }        &   0.05 \\ \hline
TC-Driver  &    0.27   &   {\bf 0.04 }    \\ \hline
\end{tabular}}
\end{table}

\section{CONCLUSION}
We presented TC-Driver, a hierarchical approach to autonomous racing, using \gls{rl} to track trajectories generated by a traditional high-level planner. 
Given imperfect modeling of parameters, \gls{mpc}'s optimality does not hold, leading to slower lap times and potentially even crashes. \gls{rl} offers a viable approach to this solution by generalizing to different driving conditions. Yet, 
end-to-end \gls{rl} methods, rely on states that are not fit for efficient generalization to different tracks and model mismatch. 
Combining a traditionally generated trajectory as an observation for a \gls{rl} agent tracking the trajectory under changing conditions alleviates these shortcomings. We evaluated and compared these approaches in the simulated F1TENTH autonomous racing environment \cite{okelly2020f1tenth}. The proposed TC-Driver shows that it can adapt to model mismatch scenarios that the non-learning based \gls{mpc}, based on \cite{liniger_mpcc} fails to handle. Further, it outperforms the model-free end-to-end architectures based on \cite{chisari2021learning, fuchs2021, song_autonomous_2021}, in all metrics regarding the robustness to model mismatch. It achieves lower and more consistent lap times, compared to the end-to-end agent, and has the lowest overall crash ratio in the model mismatch setting. 

Future work on this topic is the deployment and comparison of the TC-Driver on the physical F1TENTH system. This allows us to further investigate and evaluate the model mismatch robustness as well as the sim-to-real capabilities. Lastly, a highly interesting \gls{rl} approach would be the utilization of a model-based \gls{rl} architecture, as inspired by \cite{brunnbauer_model-based_2021}.

The code for reproducing all mentioned \gls{rl} and \gls{mpcc} F1TENTH implementations is available at: \href{}{https://github.com/ETH-PBL/TC-Driver}.

\section*{ACKNOWLEDGMENT}
We would like to thank Dr. Christian Vogt, Dr. Andrea Carron and Dr. Alexander Liniger of ETH Zürich, for their constructive and fruitful algorithmic discussions.  

\bibliographystyle{IEEEtran}
\bibliography{ref}

\end{document}

%% file: acr/acr.tex
\newacronym{mpc}{MPC}{Model Predictive Control}
\newacronym{mpcc}{MPCC}{Model Predictive Contouring Controller}
\newacronym{rl}{RL}{Reinforcement Learning}
\newacronym{mlp}{MLP}{Multilayer Perceptron}
\newacronym{forl}{FoRL}{Foundations of Reinforcement Learning}
\newacronym{ml}{ML}{Machine Learning}
\newacronym{sb3}{SB3}{Stable Baselines 3}
\newacronym{sac}{SAC}{Soft Actor Critic}
\newacronym{ppo}{PPO}{Proximal Policy Optimization}
\newacronym{ai}{AI}{Artificial Intelligence}
\newacronym{nn}{NN}{Neural Network}
\newacronym{sota}{SotA}{State-of-the-Art}